\RequirePackage{fix-cm}
\documentclass[twocolumn]{svjour3}

\smartqed
\usepackage{graphicx}
\usepackage{tabularx}
\usepackage{multirow}
\usepackage{cite}
\usepackage[outdir=./]{epstopdf}

\newcolumntype{Y}{>{\centering\arraybackslash}X}
\journalname{Neural Computing and Applications}

\begin{document}

  \title{On Usage of Autoencoders and Siamese Networks for Online Handwritten Signature Verification}
  \author{Kian Ahrabian \and Bagher Babaali}

  \institute{
    Kian Ahrabian \at School of Mathematics, Statistics, and Computer Science, University of Tehran, Tehran, Iran \\ Tel.: +98-9125482934 \\ \email{kahrabian@ut.ac.ir} \and
    Bagher Babaali \at School of Mathematics, Statistics, and Computer Science, University of Tehran, Tehran, Iran \\ Tel.: +98-9125248895 \\ \email{babaali@ut.ac.ir} \and
  }

  \date{Submitted: December 2017}

  \maketitle

  \begin{abstract}
    In this paper, we propose a novel writer-independent global feature extraction framework for the task of automatic signature verification which aims to make robust systems for automatically distinguishing negative and positive samples. Our method consists of an autoencoder for modeling the sample space into a fixed length latent space and a Siamese Network for classifying the fixed-length samples obtained from the autoencoder based on the reference samples of a subject as being “Genuine” or “Forged.” During our experiments, usage of Attention Mechanism and applying Downsampling significantly improved the accuracy of the proposed framework. We evaluated our proposed framework using SigWiComp2013 Japanese and GPDSsyntheticOnLineOffLineSignature datasets. On the SigWiComp2013 Japanese dataset, we achieved 8.65\% EER\footnote{Equal Error Rate} that means 1.2\% relative improvement compared to the best-reported result. Furthermore, on the GPDSsyntheticOnLineOffLineSignature dataset, we achieved average EERs of 0.13\%, 0.12\%, 0.21\% and 0.25\% respectively for 150, 300, 1000 and 2000 test subjects which indicates improvement of relative EER on the best-reported result by 95.67\%, 95.26\%, 92.9\% and 91.52\% respectively. Apart from the accuracy gain, because of the nature of our proposed framework which is based on neural networks and consequently is as simple as some consecutive matrix multiplications, it has less computational cost than conventional methods such as DTW\footnote{Dynamic Time Warping} and could be used concurrently on devices such as GPU\footnote{Graphics Processing Unit}, TPU\footnote{Tensor Processing Unit}, etc.
  \end{abstract}

  \section{Introduction}
    \label{sec:1}
    The handwritten signature is the most widely accepted biometric measure for authentication in governments, legal systems, banks, etc. Therefore it is of the utmost importance to have robust methods for verifying the identity of people based on their signatures. The task of ASV\footnote{Automatic Signature Verification} aims to address this problem by making robust systems that automatically classify a signature sample as being “Genuine” or “Forged” compared to some previously acquired signatures from the claimed subject. By doing so, we might be able to verify the identity of a person at a speed and accuracy which exceeds the human performance.

    In online signature verification, the signatures are acquired directly from a capturing device that records the dynamic information of the pen at a constant rate which usually includes coordinates, pressure, velocity, azimuth, etc. Consequently creating a sequence of data points that might also embed some other hidden dynamic information of the signature in itself too. These obscure pieces of information could be further exploited to discriminate the positive and negative samples. This method is in contrast with offline signature verification in which we acquire the sample after the signing process is complete, leaving us with only a picture of the signature. Due to saving the dynamic information of the sample instead of saving just an image, it becomes more difficult to forge a signature while trying to simulate these features. Hence the online methods for signature verification are considered to be more reliable and accurate than the offline forms.

    The previous studies on online signature verification could be grouped into two main categories based on their approaches for feature extraction \cite{40}:
    \begin{itemize}
      \item \textbf{\textit{Global Feature Extraction methods:}} The primary focus of these methods is to derive a fixed length feature vector from the signature as a whole to make them comparable. We divide this category into two subcategories. The first subcategory consists of algorithms that try to extract the features from the totality of signature. For example, in \cite{1} the number of strokes is used as a global feature. There also exist many other characteristics such as average velocity, average pressure, and the number of times the pen is lifted during the signing \cite{2}. An excellent example of the extent of these features could be found in \cite{3}, in which 100 global elements have been sorted by their discrimination power, a subset of these features has been used in other studies \cite{4, 5, 6, 7, 8}. The second subcategory consists of algorithms that obtain a fixed length feature vector of elements by applying a transformation on the signature. For example, in \cite{9} a wavelet transform has been used to extract a feature vector from the entire sample. Other study \cite{10}, use DCT\footnote{Discrete Cosine Transform} transform to obtain the fixed length feature vector \cite{10}. Also, in a recent work \cite{40}, a fixed length vector, called i-vector, is extracted from each signature sample. The low-dimensional fixed length i-vector representation was originally proposed for speaker verification, and in \cite{40} it has been adopted for the application of online signature verification.
      \item \textbf{\textit{Functional methods:}} These methods are more focused on calculating a distance between two signatures which are represented by a sequence of extracted features (Data Points) and comparing them based on the estimated distance. Like the previous one, this category also could be divided into two subcategories. In the first subcategory, the algorithms do not perform any modeling, and these methods keep a reference set for each subject and use it in the test time to classify the input signature by comparing it to reference samples. DTW method is the most well-known method in this subcategory, and it has been widely used in many studies \cite{11, 12, 13, 14}. In the second subcategory, the algorithms train a probabilistic model using the signatures in the reference set and later use this model to classify the test signatures as being forged or genuine by calculating the probability of them being genuine and comparing it with a threshold. These methods usually use likelihoods for scoring and decision making. HMM\footnote{Hidden Markov Model} \cite{15, 16, 17, 18, 19} and GMM\footnote{Gaussian Mixture Model} \cite{20, 21, 22} are the most common methods in this subcategory.
    \end{itemize}

    From another perspective, it is also possible to divide previous studies on online signature verification into two main groups based on their approach to training the classifier \cite{41}. In a Writer-Dependent method, for each subject, a binary classifier is trained based on a training set which only consists of signatures from the specific subject. In a Writer-Independent approach, a single global classifier is trained based on the whole training set. In both methods, the negative samples are obtained from the forgery samples of the specific subject which is provided in the training set. The main benefit of a writer-independent approach is that it needs much less computation power compared to the other method. Also, because these methods do not require to train a new model for each new user, they are considered easier to deploy and use in different scenarios. Note that it is also possible to use other subjects signatures as negative samples, but due to the nature of this method, it could lead to poor results and degraded generalization ability. At the test time, we use a classifier which is trained with one of the mentioned approaches to classify the query signature as being “Forged” or “Genuine” compared to reference signatures of the claimed individual. Some comprehensive reviews on the problem could be found in \cite{23} and \cite{24}.

    In this paper, we propose a Writer-Independent Global Feature Extraction method which is based on a Recurrent Autoencoder, sometimes called a Diabolo Network \cite{25}, as the global feature extractor and a Siamese Network as the global classifier. Our approach for training the Recurrent Autoencoder is based on the proposed technique in \cite{26} which is called Sequence to Sequence Learning. In this method, a model is trained to learn a mapping between samples space where every sample is a sequence of data points and a fixed length latent space. By doing so, we obtain a mapping that could be used to extract a fixed length feature vector that captures the most crucial attributes of each sample. We expect that these feature vectors could be further used to distinguish the forged and genuine samples. As for the classification part, we train a Siamese Neural Network. Characteristically, siamese networks are a perfect solution for verification tasks, as has been previously studied in \cite{27, 28, 29, 30, 31}, because their objective is to learn a discriminative similarity measure between two or more inputs that classifies the inputs as being the same or not. So we expect to achieve a promising solution with this combination for the task of Online Signature Verification.

    It is important to note that the proposed method has a relatively low cost of computation which is mostly during the training time and it reduces to just a few matrix multiplications at the test time. This is a crucial advantage because it could lead to faster verification systems which can respond quickly to the queries by concurrently computing the results on devices such as GPU, TPU, etc. 

    We organize the rest of the paper as follows: In Sect. \ref{sec:2} we describe the architecture and attributes of the models we used along with a detailed description of their structure. Sect. \ref{sec:3} and Sect. \ref{sec:4} presents our experiments and its comparison to the state-of-the-art algorithms. Finally, in Sect. \ref{sec:5}, we conclude our paper with possible future directions.

  \section{The Proposed Framework}
    \label{sec:2}
    In this section, at first, the preprocessing performed on samples is explained in Sect. \ref{sec:2.1}. This is followed by an overview of the proposed Autoencoder architecture in Sect. \ref{sec:2.2}. Finally, a detailed description of the proposed Siamese Network architecture is given in Sect. \ref{sec:2.3}.

    \subsection{Preprocessing}
      \label{sec:2.1}
      At first, we extract a set of 12 local features from the provided coordinates of the pen for each data point. These features are among the local features that have been introduced in \cite{7, 40}. These locally extracted features are listed in Table \ref{tab:1}. After obtaining the local features, we normalize each sample individually by applying the Standard Normalization method on each local feature of each sample independently. Then a threshold is applied to the length of samples to filter out very long samples.

      \begin{table}
        \caption{List of extracted local features for each data point.}
        \label{tab:1}
        \begin{tabularx}{\linewidth}{|X|X|}
          \hline 1. Horizontal position $x_t$ & 2. Vertical position $y_t$ \\
          \hline 3. Path-tangent angle & 4. Path velocity magnitude \\
          \hline 5. Log curvature radius & 6. Total acceleration magnitude \\
          \hline \multicolumn{2}{|c|}{7-12. First order derivatives of 1-6} \\
          \hline
        \end{tabularx}
      \end{table}

      At this stage, we apply a downsampling procedure to each sample. To perform this process with a sampling rate of $K$, we traverse the data points concerning the chronological order and do the following steps repeatedly until we reach the end of the sequence:
      \begin{enumerate}
        \item Select a data point
        \item Skip $K - 1$ data points
      \end{enumerate}
      By applying this method to a given sample with the length of $L$, we obtain a downsampled sample with the length of $[\frac{L}{K}]$.

      Finally, since batch training in neural networks requires samples to possess the same shape, we must apply a padding procedure to obtain samples with equal length. To do this, for each sample, we use pre-padding with zero to produce a sample that has a length equal to the length of the most extended sample in the dataset.

    \subsection{Autoencoder}
      \label{sec:2.2}
      The RNN\footnote{Recurrent Neural Network} is a generalization of feedforward neural networks that make it possible to process sequences sequentially by maintaining a hidden state $h$ which works as a memory containing the previously observed context. Given a sequence of inputs $(x_1, ..., x_t)$, a standard RNN generates a sequence of outputs $(y_1, ..., y_t)$ in which the output at timestep $t$ is computed by
      \begin{equation}
        h_t = f(h_{t-1}, x_t),
      \end{equation}
      \begin{equation}
        y_t = W_hh_t,
      \end{equation}
      where $f$ is a non-linear activation function \cite{26}.

      It is possible to train an RNN to learn a probability distribution over a set of sequences to predict the next symbol in a sequence given the previous symbols. By doing so, the output at each time step $t$ could be formulated as the conditional distribution $p(x_t | x_{t−1}, ..., x_1)$. Thus the probability of the sequence $x$ is calculated using
      \begin{equation}
        p(x) = \prod_{t=1}^{T} p(x_t | x_{t-1}, ..., x_1).
      \end{equation}
      After learning the distribution, we could iteratively predict a symbol at each time step to obtain a whole sequence from a sample in the learned distribution \cite{39}.
      
      BRNN\footnote{Bidirectional Recurrent Neural Networks} \cite{32} is an extension to the vanilla RNN. The idea behind these networks is to have two separate RNN and feed each training sequence forward and backward to them, note that both of these RNNs are connected to the same output layer. At each point in a given sequence, this structure provides the network with information about all points before and after it, and this means that the BRNN has complete sequential details of the sequence in each step of the computation. Online signature verification, along with some other “online” tasks, requires an output after the end of the whole sequence. Therefore it is plausible to use BRNNs to gain improved performance on these tasks.

      The main difficulty of training RNNs is to learn a model that captures the long-term dependencies. This problem usually appears when we are dealing with long sequences that have dependencies between distant data points. Intuitively it seems that the RNNs are unable to remember the information that is embedded in previously visited distant data points. In theory, RNNs are entirely capable of handling this problem, but unfortunately, in practice, RNNs don’t seem to be able to learn them \cite{33, 34}. To address this issue, we use a particular kind of RNN called LSTM\footnote{Long Short-Term Memory} \cite{35}. These networks are explicitly designed to deal with the long-term dependency problem and have proven their capability of learning long-term dependencies by showing excellent performance on a large variety of problems.

      Since the length of signature samples could be as high as thousands of data points, even with the use of LSTM networks the problem of long-term dependencies remains. To improve the performance we use a technique called Attention Mechanism \cite{36, 37, 38}. Intuitively the model tries to learn what to attend to based on the input sequence and the output sequence of the network. By doing so, each output $y_t$ of the decoder depends on a weighted combination of all the input states instead of only relying on the last hidden state of the network. Beside the attention mechanism, to improve the performance of the model, we feed the input backward to the encoder. This trick has been introduced in \cite{39} and has shown a positive effect on the performance of the model by introducing many short-term dependencies, making the optimization problem more straightforward to solve.

      Because of their structure, LSTM networks are more prone to overfitting which is a result of memorizing the training set samples. This problem will lead to a degraded performance and the inability of the model to generalize to new signatures. We use a regularization method called Dropout which has been introduced in \cite{42, 43} to overcome this problem. The intuition behind this approach is that if we randomly drop units of the network with a probability of $p$ in the training phase, creating a somewhat sparse network, we can prevent the co-adapting of the units and as a result obtaining a model with good generalization ability.

\begin{figure*}
        \begin{center}
          \includegraphics[width=\textwidth]{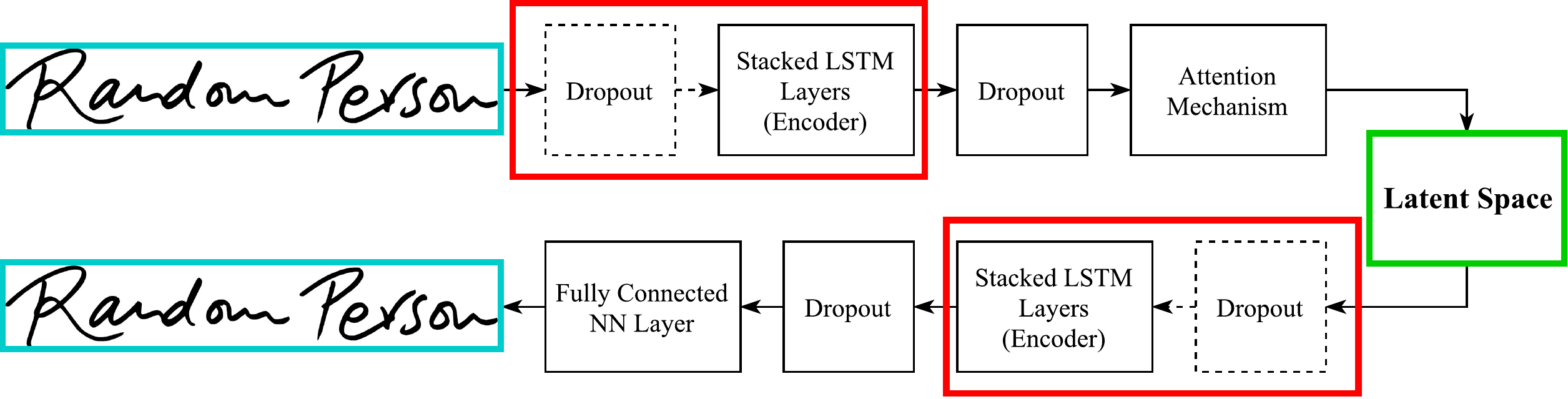}
          \caption{Autoencoder architecture. The red rectangles show the parts that could be repeated. The dots boxes show the optional components.}
          \label{fig:1}
        \end{center}
      \end{figure*}

      We used an Autoencoder architecture that closely follows the work of \cite{26} except applying Bidirectional LSTM with Attention Mechanism instead of Vanilla RNN because of the previously mentioned benefits in boosting the performance. This architecture is illustrated in Fig. \ref{fig:1}. As for the other parameters, we used a combination of Glorot Uniform initializer function \cite{44}, ReLU\footnote{Rectified Linear Units} activation function \cite{45}, ADAM optimizer \cite{46}, Mini Batch Size of 128, MAPE\footnote{Mean Absolute Percentage Error} loss function and Early Stopping on loss value \cite{47} to achieve improved performance and training speed. The details of our configuration are illustrated in Table \ref{tab:2}. For the training phase, we used the preprocessed sequences which were obtained by the previously described method from the raw coordinate samples.

      The encoder part of the trained autoencoder will be further used to map each of the given sequences to the fixed-length vector space.

      \begin{table}
        \caption{Detailed information of configuration used to train autoencoder.}
        \label{tab:2}
        \begin{tabularx}{\linewidth}{|Y|Y|}
          \hline \textbf{Attribute} & \textbf{Used Value} \\
          \hline Initializer Function & Glorot Uniform \\
          \hline Activation Function & ReLU \\
          \hline Mini Batch Size & 128 \\
          \hline Loss Function & MAPE \\
          \hline \multirow{3}{*}{Optimizer} & \multirow{3}{\linewidth}{\centering ADAM, $\alpha = 0.001$, $\beta_1 = 0.9$, $\beta_2 = 0.999$, $\epsilon = 1\mathrm{e}{-8}$} \\ & \\ & \\
          \hline Early Stopping & $Patience = 10$, $Min \Delta = 0$ \\
          \hline Epochs & 1000 \\
          \hline
        \end{tabularx}
      \end{table}

    \subsection{Siamese Network}
      \label{sec:2.3}
      Siamese networks are a particular kind of neural networks that are made of two input fields followed by two identical neural network called legs with shared weights. These two-legged architecture are eventually merged into a single layer by applying a discriminative function to the resulting output of the legs. These networks are designed to compare two inputs, and as a result, they are a reasonable pick when it comes to verification tasks.

\begin{figure*}
        \begin{center}
          \includegraphics[width=\textwidth]{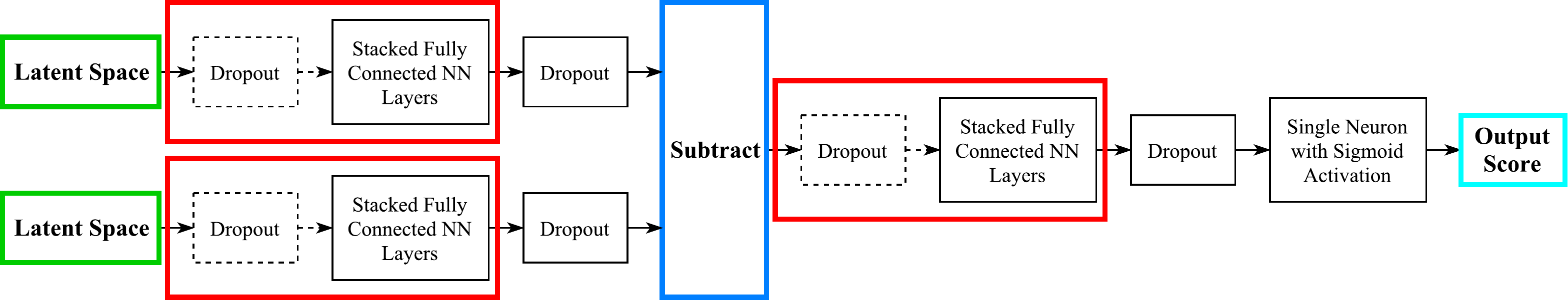}
          \caption{Siamese network architecture. The red rectangles show the parts that could be repeated. The dots boxes show the optional components.}
          \label{fig:2}
        \end{center}
      \end{figure*}

      We used a Siamese Network which consisted of one fully connected layer before and after the merge procedure followed by a single neuron with sigmoid activation function. The goal of this network is to calculate the probability of two inputs being in the same class. This network enables us to make a decision in the classification step based on the calculated probabilities concerning different reference samples which makes our classifier more accurate than the classic direct classification method. This architecture is depicted in Fig. \ref{fig:2}.

      To prevent the network from overfitting, we used the Dropout method for regularization. As for the other parameters, similar to the autoencoder, we used a combination of Glorot Uniform initializer function, ReLU activation function, ADAM optimizer, Mini Batch Size of 128, Binary Cross Entropy loss function and Early Stopping on loss value to achieve improved performance and training speed. The details of our configuration are illustrated in Table \ref{tab:3}. In training phase, we used the fixed length vectors obtained from the encoder part of the previously described autoencoder. A mean normalization method was applied before feeding the samples to the network by calculating a mean vector from the feature vectors of reference signatures of each subject and subtracting it from the feature vectors of each signature of the corresponding subject independently.

      \begin{table}
        \caption{Detailed information of configuration used to train siamese network.}
        \label{tab:3}
        \begin{tabularx}{\linewidth}{|Y|Y|}
          \hline \textbf{Attribute} & \textbf{Used Value} \\
          \hline Initializer Function & Glorot Uniform \\
          \hline Activation Function & ReLU \\
          \hline Mini Batch Size & 128 \\
          \hline Loss Function & Binary Cross Entropy \\
          \hline \multirow{3}{*}{Optimizer} & \multirow{3}{\linewidth}{\centering ADAM, $\alpha = 0.001$, $\beta_1 = 0.9$, $\beta_2 = 0.999$, $\epsilon = 1\mathrm{e}{-8}$} \\ & \\ & \\
          \hline Early Stopping & $Patience = 10$, $Min \Delta = 0$ \\
          \hline Epochs & 1000 \\
          \hline
        \end{tabularx}
      \end{table}

\begin{figure*}
        \begin{center}
          \includegraphics[scale=0.5]{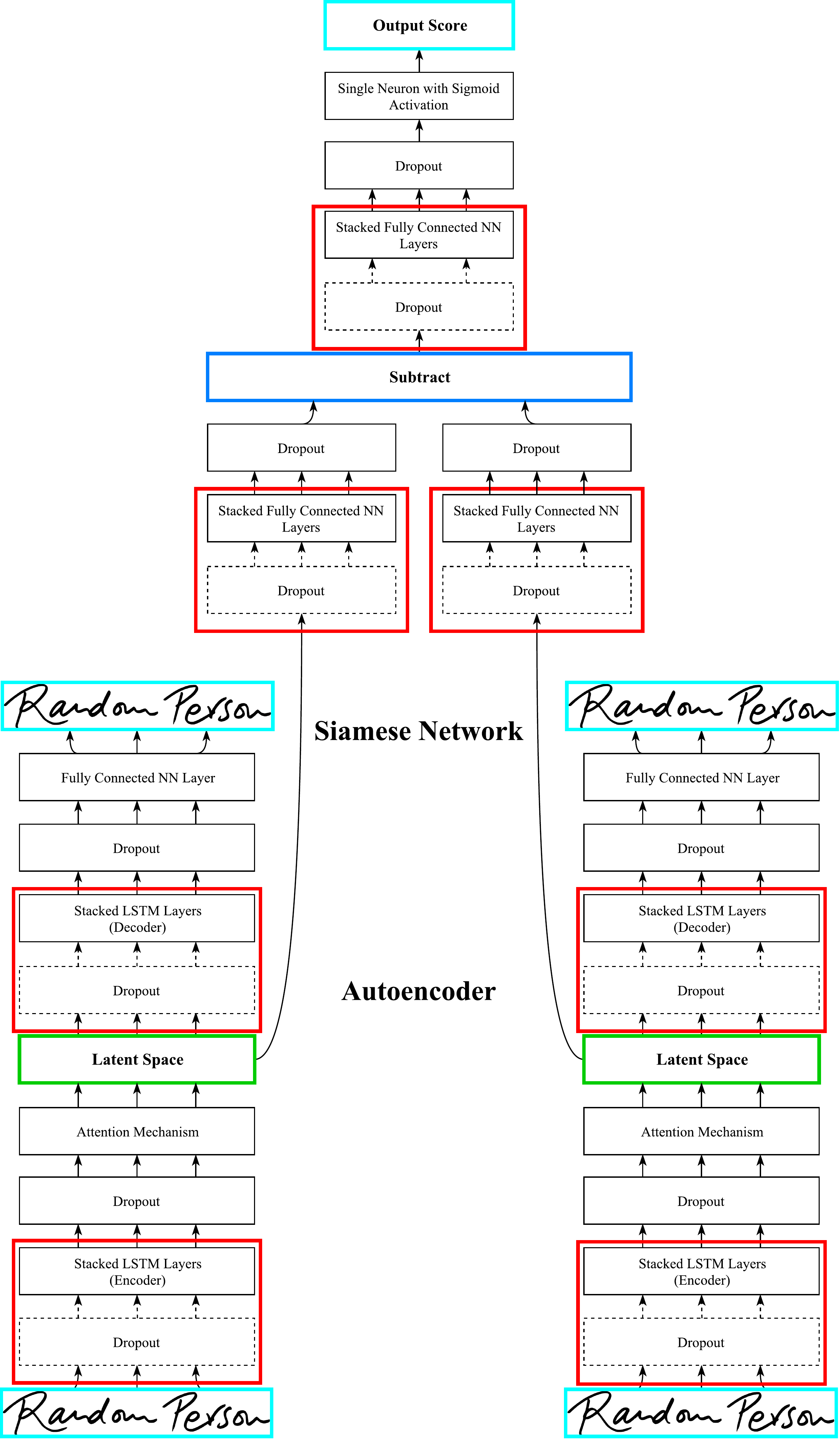}
          \caption{The complete pipeline of our proposed method. The red rectangles show the parts that could be repeated. The dots boxes show the optional components.}
          \label{fig:3}
        \end{center}
      \end{figure*}

      In the test phase, first, we feed the extracted feature vector of the given sample with every extracted feature vector of the reference samples of the claimed individual to the trained siamese network and calculate the outputs which are probabilities of them being in the same class. Then we apply a predefined threshold to the estimated probabilities to classify them as being in the same category or not. Eventually, we classify a sample as “Genuine” if it was classified as being in the same class with more than a predefined number of the reference samples of the claimed individual. Otherwise, we classify the sample as being “Forged.” A complete illustration of our proposed framework is shown in Fig. \ref{fig:3}.

  \section{Experiments Setup}
    \label{sec:3}
    For evaluation we used the SigWiComp2013 Japanese \cite{48} and GPDSsyntheticOnLineOffLineSignature \cite{49} datasets. Our codebase is entirely implemented using Keras \cite{53} library with Theano \cite{52} as the backend. We trained the models using a K40 GPU, and it approximately took 1 to 24 hours to finish the training, depending on different datasets and different training settings.

    \subsection{Datasets}
      \label{sec:3.1}
      In this section, we describe the attributes of the datasets we used for the evaluation.

      \subsubsection{SigWiComp2013 Japanese}
        \label{sec:3.1.1}
        The SigWiComp2013 Japanese dataset contains the signatures of 31 subjects, 11 for the training set and 20 for the test set. For each subject, 42 genuine and 36 forged samples are available. This results in 462 genuine and 396 forged signatures in the training set and 840 genuine and 720 forged signatures in the test set. Each subject in the test set has a set of twelve reference signatures. The exact description of autoencoder and siamese network used for this dataset are illustrated in Table \ref{tab:4} and Table \ref{tab:5} respectively.

        \begin{table}
          \caption{Detailed information on Autoencoder structure for SigWiComp2013 Japanese dataset.}
          \label{tab:4}
          \begin{tabularx}{\linewidth}{|Y|Y|}
            \hline \textbf{Layer} & \textbf{Description} \\
            \hline Bidirectional LSTM (Encoder) & 64 units for each direction \\
            \hline Dropout & With probability of 0.5 \\
            \hline Attention Mechanism & Follows the work of \cite{37} \\
            \hline Bidirectional LSTM (Decoder) & 64 units for each direction \\
            \hline Dropout & With probability of 0.5 \\
            \hline Dense & 12 units \\
            \hline
          \end{tabularx}
        \end{table}

        \begin{table}
          \caption{Detailed information on Siamese Network structure for SigWiComp2013 Japanese dataset.}
          \label{tab:5}
          \begin{tabularx}{\linewidth}{|Y|Y|}
            \hline \textbf{Layer} & \textbf{Description} \\
            \hline Dense (Before Merge) & 128 units \\
            \hline Dropout & With probability of 0.5 \\
            \hline Merge & Subtract the output of the second leg from the output of the first leg \\
            \hline Dense (Before Merge) & 128 units \\
            \hline Dropout & With probability of 0.5 \\
            \hline Dense & One unit with sigmoid activation \\
            \hline
          \end{tabularx}
        \end{table}

      \subsubsection{GPDSsyntheticOnLineOffLineSignature}
        \label{sec:3.1.2}
        The GPDSsyntheticOnLineOffLineSignature dataset was built based on the procedure in \cite{49}. It contains the signatures of 10000 subjects. For each subject, 24 genuine and 30 forged signatures are available. For evaluation, we used two non-intersecting subsets of signatures for the training set and the test set with different sizes combination of 150, 300, 1000 and 2000 subjects. Each subject in the test set had a set of five reference signatures. The exact architectures of autoencoder and siamese network used for this network are illustrated in Table \ref{tab:6} and Table \ref{tab:7} respectively.

        \begin{table}
          \caption{Detailed information on Autoencoder structure for GPDSsyntheticOnLineOffLineSignature dataset.}
          \label{tab:6}
          \begin{tabularx}{\linewidth}{|Y|Y|}
            \hline \textbf{Layer} & \textbf{Description} \\
            \hline Bidirectional LSTM (Encoder) & 64 units for each direction \\
            \hline Dropout & With probability of 0.65 \\
            \hline Bidirectional LSTM (Encoder) & 64 units for each direction \\
            \hline Dropout & With probability of 0.65 \\
            \hline Attention Mechanism & Follows the work of \cite{37} \\
            \hline Bidirectional LSTM (Decoder) & 64 units for each direction \\
            \hline Dropout & With probability of 0.65 \\
            \hline Bidirectional LSTM (Decoder) & 64 units for each direction \\
            \hline Dropout & With probability of 0.65 \\
            \hline Dense & 12 units \\
            \hline
          \end{tabularx}
        \end{table}

        \begin{table}
          \caption{Detailed information on Siamese Network structure for GPDSsyntheticOnLineOffLineSignature dataset.}
          \label{tab:7}
          \begin{tabularx}{\linewidth}{|Y|Y|}
            \hline \textbf{Layer} & \textbf{Description} \\
            \hline Dense (Before Merge) & 128 units \\
            \hline Dropout & With probability of 0.5 \\
            \hline Merge & Subtract the output of the second leg from the output of the first leg \\
            \hline Dense (Before Merge) & 128 units \\
            \hline Dropout & With probability of 0.5 \\
            \hline Dense & One unit with sigmoid activation \\
            \hline
          \end{tabularx}
        \end{table}

      \subsection{Experiment Scenarios}
        \label{sec:3.2}
        We evaluated our proposed method with two somewhat different scenarios. In the first one, for training the Autoencoder we used every available sample in the training set, forged and genuine, along with the reference samples from the test set. For the Siamese Network, we used the cross product of forged and genuine feature vectors of each subject in the training set as negative samples. As for the positive samples, we used the union of the following sets of elements:
        \begin{enumerate}
          \item Every combination of genuine signatures of each subject in the training set
          \item Every combination of forged signatures of each subject in the training set
          \item Every combination of reference signatures of each subject in the test set
        \end{enumerate}

        In the second scenario, to train the Autoencoder, we only used the available samples in the training set and dropped the reference samples from the test set. For the Siamese Network, similar to the first scenario, we used the cross product of forged and genuine feature vectors of each subject in the training set as negative samples. As for the positive samples, we used the union of the following sets of elements:
        \begin{enumerate}
          \item Every combination of genuine signatures of each subject in the training set
          \item Every combination of forged signatures of each subject in the training set
        \end{enumerate}

        It is essential to discuss the plausibility of using the reference samples in the test set for the training phase. If we consider a real-world scenario, in each of the previously mentioned use-cases every subject has some reference signatures which has been taken for future references. Hence in a real-world situation, we have access to the reference signatures of each subject at any time. As a result, it is entirely plausible to use the reference signatures of the test set during the training phase.

        Another relevant discussion that could be argued is about using skilled forgery samples provided in training set for the training process. Since our proposed framework consists of two parts in which the second part, siamese network, need negative samples as well as positive samples in the trained phase, it is essential for our framework to use these negative samples for training the network. This issue could be assumed as a disadvantage compared to other methods such as DTW which does not need skilled forgery samples to be trained. These methods usually try to find a threshold for discriminating the positive and negative samples. However, these other methods could also benefit from these samples in the process of fine-tuning. These samples will allow them to achieve a more accurate threshold which leads to producing better models. All things considered, since obtaining the skilled forgery samples for a small number of subjects is a feasible possibility, this problem could be overlooked, and thus our proposed framework could also be a viable option in a real-world situation.

        \subsection{t-SNE visualization}
          \label{sec:3.3}
\begin{figure*}
          \begin{center}
            \includegraphics[width=\textwidth]{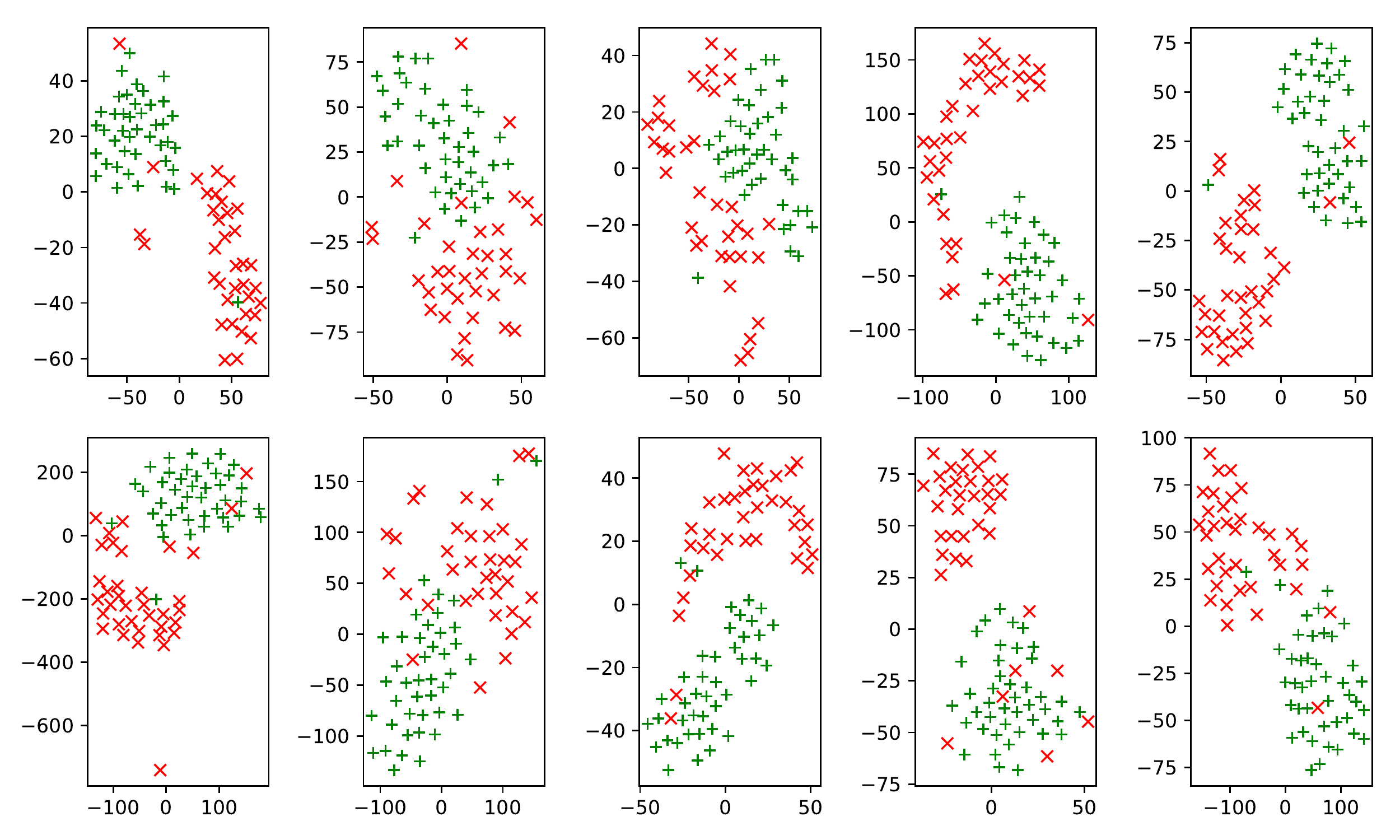}
            \caption{t-SNE plots for genuine and forged signatures for ten individuals in the test set of SigWiComp2013 Japanese dataset using the vectors obtained from the first layer of the siamese network. The red crosses show the forged signatures, and the green pluses show the genuine samples.}
            \label{fig:4}
          \end{center}
        \end{figure*}

          The best vector representation of samples for the task of signature verification is the one that flawlessly discriminates between genuine and skilled forgery samples. Fig. \ref{fig:4} shows t-SNE plots for the genuine and forged signatures of ten individuals in the test set of SigWiComp2013 Japanese dataset using the vectors obtained from the first layer of the siamese network. This representation discriminates between positive and negative samples accurately for almost all of the subjects as clearly shown in Fig. \ref{fig:4}. Therefore, we anticipate that using a suitable classifier an acceptable performance is achievable.

  \section{Experiments Results}
    \label{sec:4}
    First, we present the results obtained on SigWiComp2013 Japanese dataset in Sect. \ref{sec:4.1}. This is followed by discussions on the effect of attention mechanism and downsampling rate in Sect. \ref{sec:4.2} and Sect. \ref{sec:4.3}, respectively. Sect. \ref{sec:4.4} explains the effect of the samples lengths on the achieved result in Sect. \ref{sec:4.1}. Then we have the results acquired on GPDSsyntheticOnLineOffLineSignature dataset presented in Sect. \ref{sec:4.5}. In the end, we discuss the effect of the reference set size in Sect. \ref{sec:4.6}.
    
    \subsection{Results on SigWiComp2013 Japanese dataset}
      \label{sec:4.1}
      Table \ref{tab:8} shows the accuracies, false acceptance rates and false rejection rates of our proposed method along with the state-of-the-art techniques on previously discussed SigWiComp2013 Japanese dataset in Sect. \ref{sec:3.1}. Because of the low number of subjects in this dataset, 11 for the training set and 20 for the test set, we used the first scenario described in Sect. \ref{sec:3.2} to obtain more samples for the training phase. As shown in Table \ref{tab:8}, our method outperformed all of the previously proposed state-of-the-art methods on this dataset.

    \subsection{Effect of Attention Mechanism}
      \label{sec:4.2}
      An noticeable observation in our experiments was the effect of the attention mechanism on the performance of the proposed framework. As shown in the Table \ref{tab:8}, using this technique has improved the performance of our model significantly. This observation is aligned with other previous use cases of this method \cite{36, 37}. Intuitively, by applying attention mechanism, we end up with a model that has learned to attend to the more critical features of a given sample which is the ultimate goal of this technique.

      \begin{table*}
        \caption{Comparison of the proposed method with six other state-of-the-art methods on the SigWiComp2013 Japanese dataset \cite{40}.}
        \label{tab:8}
        \begin{tabularx}{\textwidth}{|Y|c|c|c|c|c|}
          \hline \textbf{Dataset} & \textbf{State-of-the-art Methods} & \textbf{Subjects} & \textbf{Accuracy} & \textbf{FAR} & \textbf{FRR} \\
          \hline \multirow{7}{\linewidth}{\centering SigWiComp2013 Japanese} & Qatar University & \multirow{7}{*}{\centering 31} & 70.55\% & 30.22\% & 29.56\% \\
          \cline{2-2}\cline{4-6} & Sabanci University-1 & & 72.55\% & 27.37\% & 27.56\% \\
          \cline{2-2}\cline{4-6} & Sabanci University-2 & & 72.47\% & 27.50\% & 27.56\% \\
          \cline{2-2}\cline{4-6} & i-vector + NAP-2 & & 89.06\% & 10.97\% & 10.89\% \\
          \cline{2-2}\cline{4-6} & i-vector + WCCN-2 & & 89.37\% & 10.69\% & 10.56\% \\
          \cline{2-2}\cline{4-6} & i-vector + SVM-2 & & 91.25\% & 8.75\% & 8.75\% \\
          \cline{2-2}\cline{4-6} & \textbf{Autoencoder (No Attention) + Siamese} & & \textbf{84.31\%} & \textbf{15.65\%} & \textbf{15.73\%} \\
          \cline{2-2}\cline{4-6} & \textbf{Autoencoder + Siamese} & & \textbf{91.35\%} & \textbf{8.60\%} & \textbf{8.70\%} \\
          \hline
        \end{tabularx}
      \end{table*}

    \subsection{Effect of the downsampling rate}
      \label{sec:4.3}
      Since downsampling reduces the size of a sample by a factor of $K$, it would have a significant impact on the processing speed of the framework. But the effect of this method on the performance should also be considered. Because when a downsampling is applied on a sample, many of its data points are removed, leaving us with a sort of summarization of the signature which might not have all of the information stored in the original sample. From another perspective, it is also possible to have a boost in performance by applying this method. Because by employing this technique we remove many of the data points that are very similar to other data points, making the problem easier to solve. As shown by the results in Table \ref{tab:9}, changing the downsampling rate could result in both a better or worse performance. In our experiments, a downsampling rate of five gave us the best result.

      \begin{table}
        \caption{Accuracies, false acceptance rates and false rejection rates with respect to different downsampling rates. These results are on SigWiComp2013 Japanese dataset.}
        \label{tab:9}
        \begin{tabularx}{\columnwidth}{|c|Y|Y|Y|}
          \hline \textbf{Sampling Rate} & \textbf{Accuracy} & \textbf{FAR} & \textbf{FRR} \\
          \hline 3 & 79.06\% & 20.92\% & 20.96\% \\
          \hline 4 & 90.29\% & 9.71\% & 9.71\% \\
          \hline 5 & \textbf{91.35\%} & \textbf{8.60\%} & \textbf{8.70\%} \\
          \hline 6 & 78.45\% & 21.59\% & 21.51\% \\
          \hline
        \end{tabularx}
      \end{table}

    \subsection{Effect of the samples lengths}
      \label{sec:4.4}
\begin{figure}
        \begin{center}
          \includegraphics[scale=0.5]{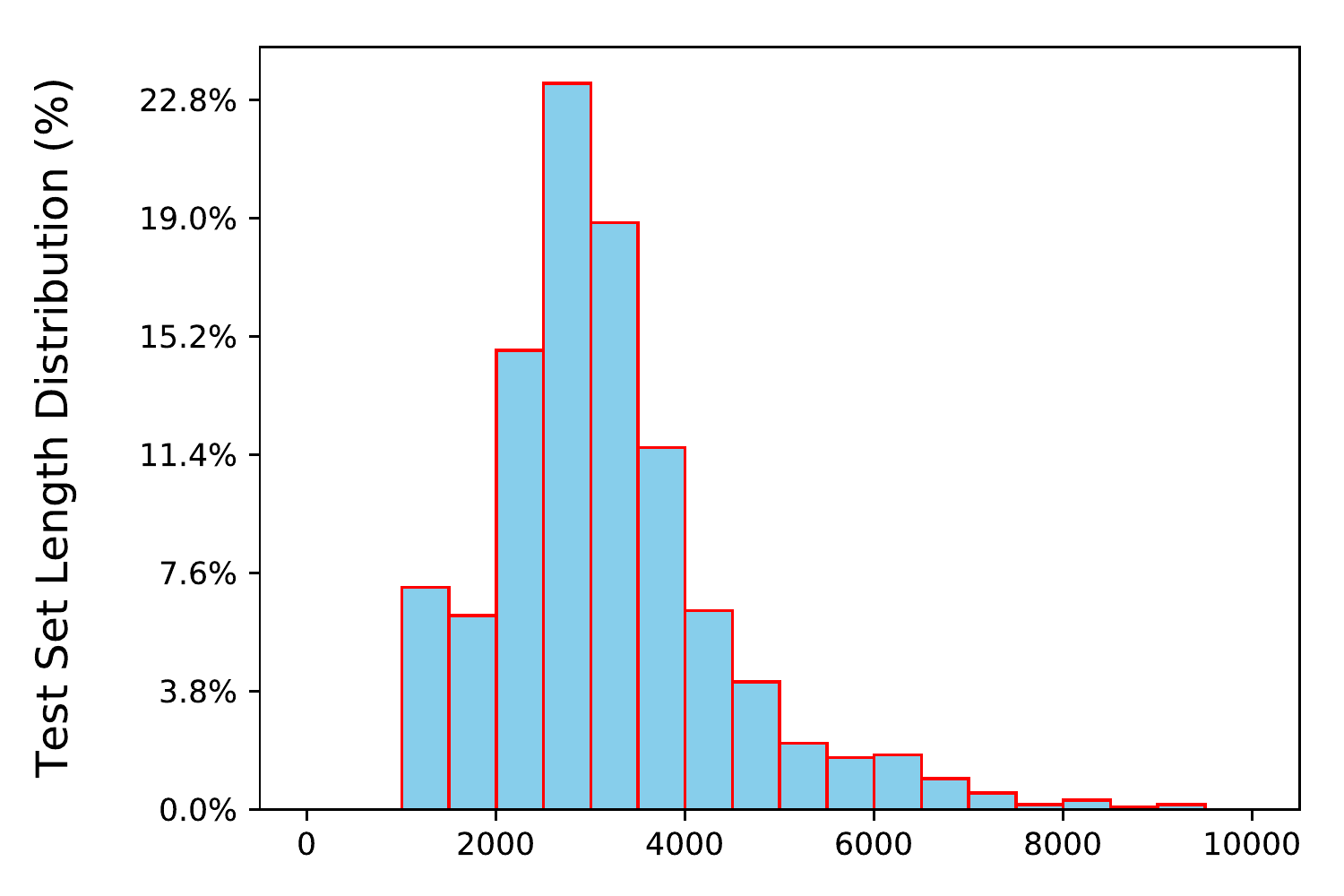}
          \caption{Length distribution of the samples in the test set.}
          \label{fig:5}
        \end{center}
      \end{figure}

\begin{figure}
        \begin{center}
          \includegraphics[scale=0.5]{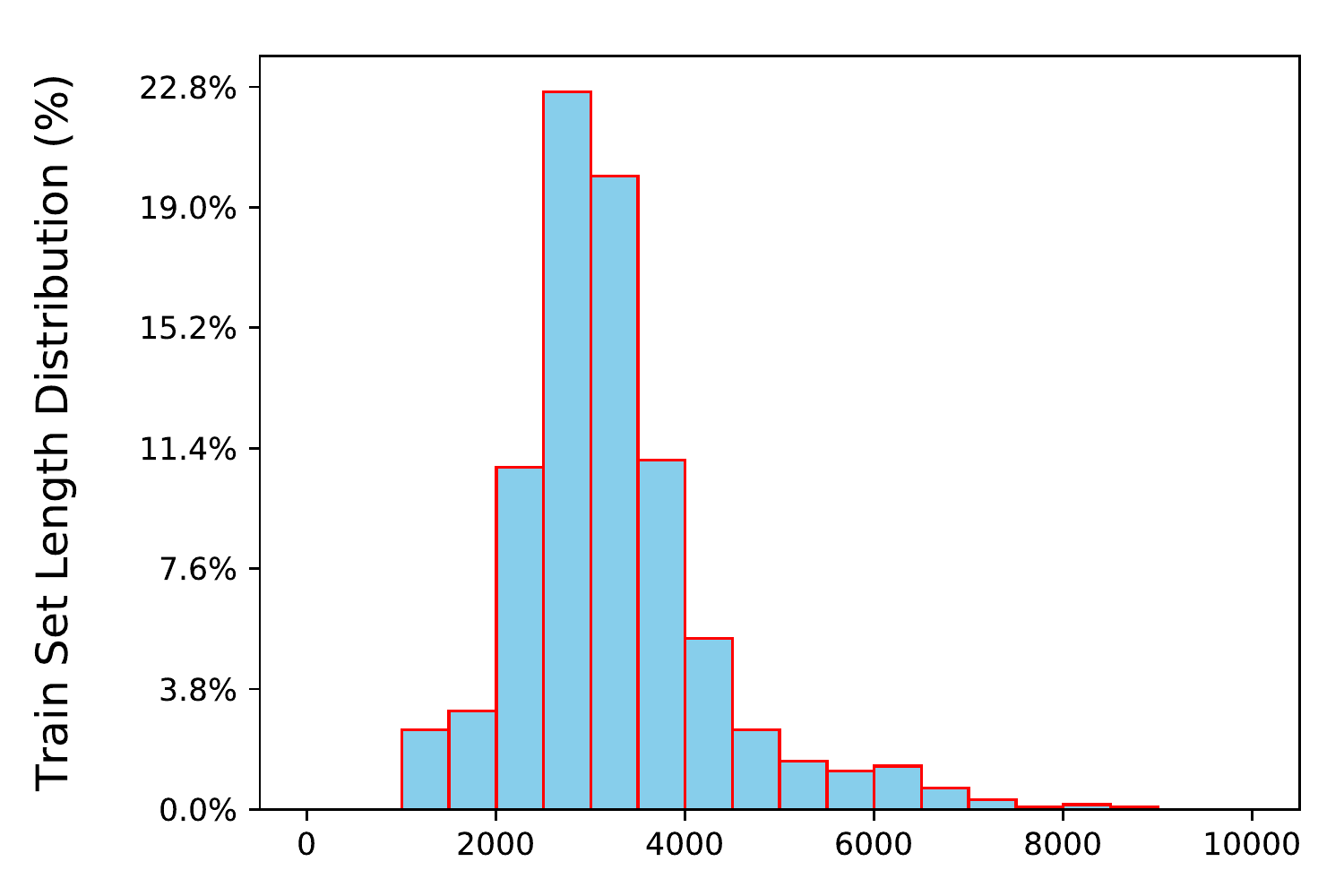}
          \caption{Length distribution of the samples in the train set.}
          \label{fig:6}
        \end{center}
      \end{figure}

\begin{figure}
        \begin{center}
          \includegraphics[scale=0.5]{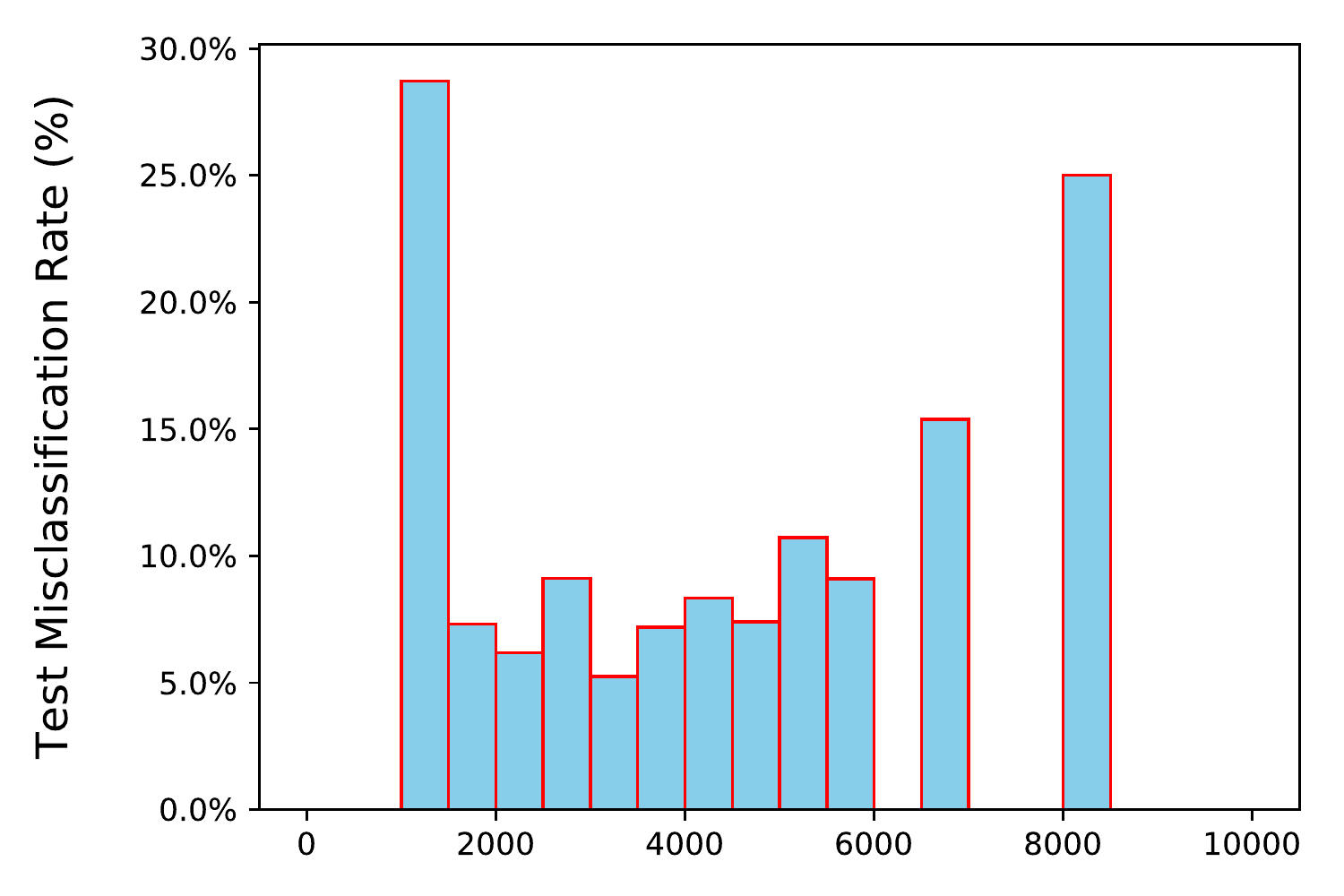}
          \caption{Relative misclassification rate of the test samples compared to the total available samples in the corresponding bin. Samples are put into separate bins based on their lengths.}
          \label{fig:7}
        \end{center}
      \end{figure}

      As demonstrated in Fig. \ref{fig:5}, Fig. \ref{fig:6} and Fig. \ref{fig:7}, the proposed framework perform well in situations that enough samples with similar length are available in the train set. Thus it can generalize well to new samples when enough samples with a proper distribution in lengths are provided in the train set. Another observation is that even with a low number of very long train samples the model has learned to model these very long samples better than the rather shorter samples. In conclusion, this framework tends to perform better in circumstances that either the samples are very long or enough samples with similar lengths are provided in the train set.

    \subsection{Results on GPDSsyntheticOnLineOffLineSignature dataset}
      \label{sec:4.5}
\begin{figure}
        \begin{center}
          \includegraphics[scale=0.5]{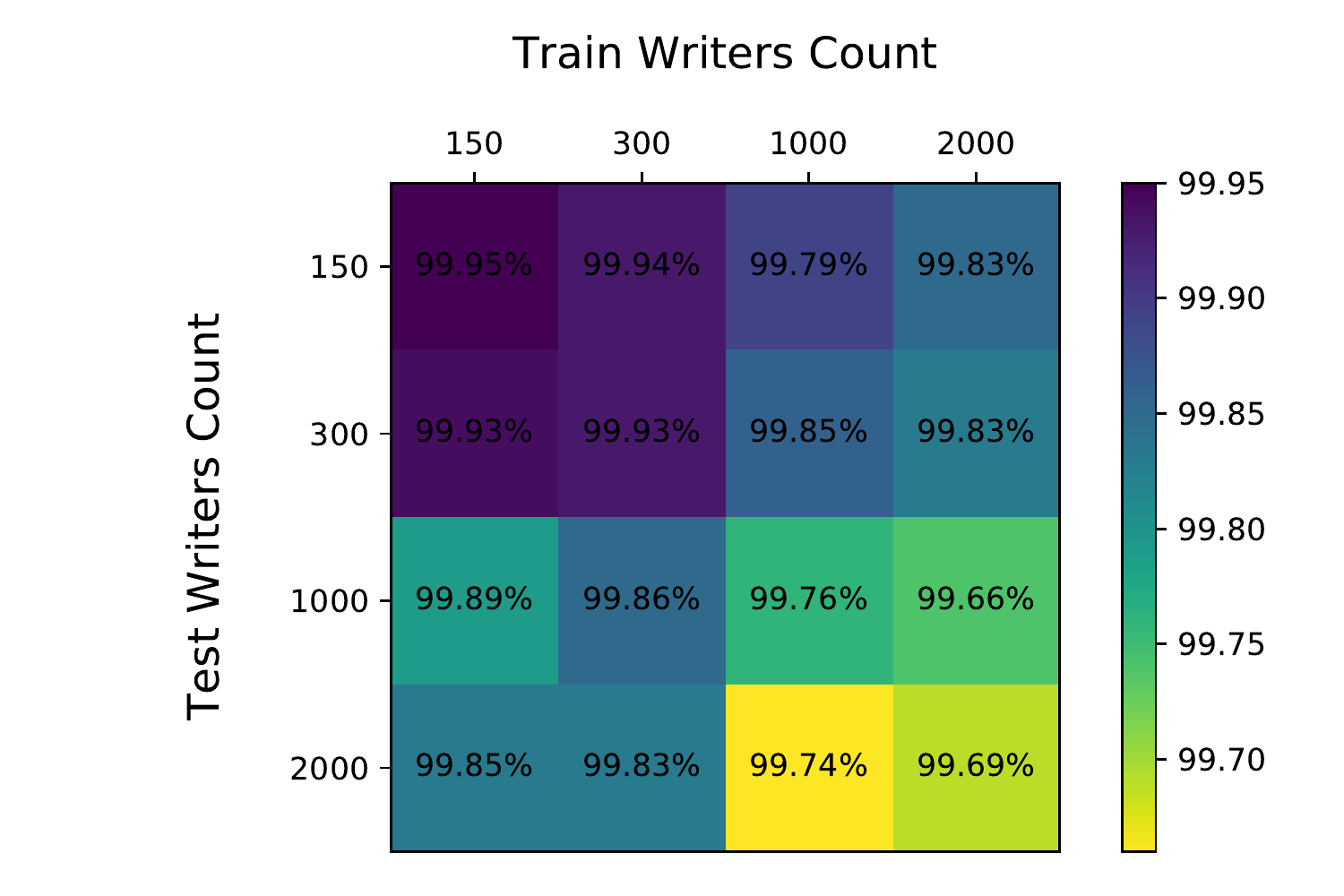}
          \caption{Accuracies of the proposed method on the GPDSsyntheticOnLineOffLineSignature dataset with different sizes of the training set and the test set.}
          \label{fig:8}
        \end{center}
      \end{figure}

\begin{figure}
        \begin{center}
          \includegraphics[scale=0.5]{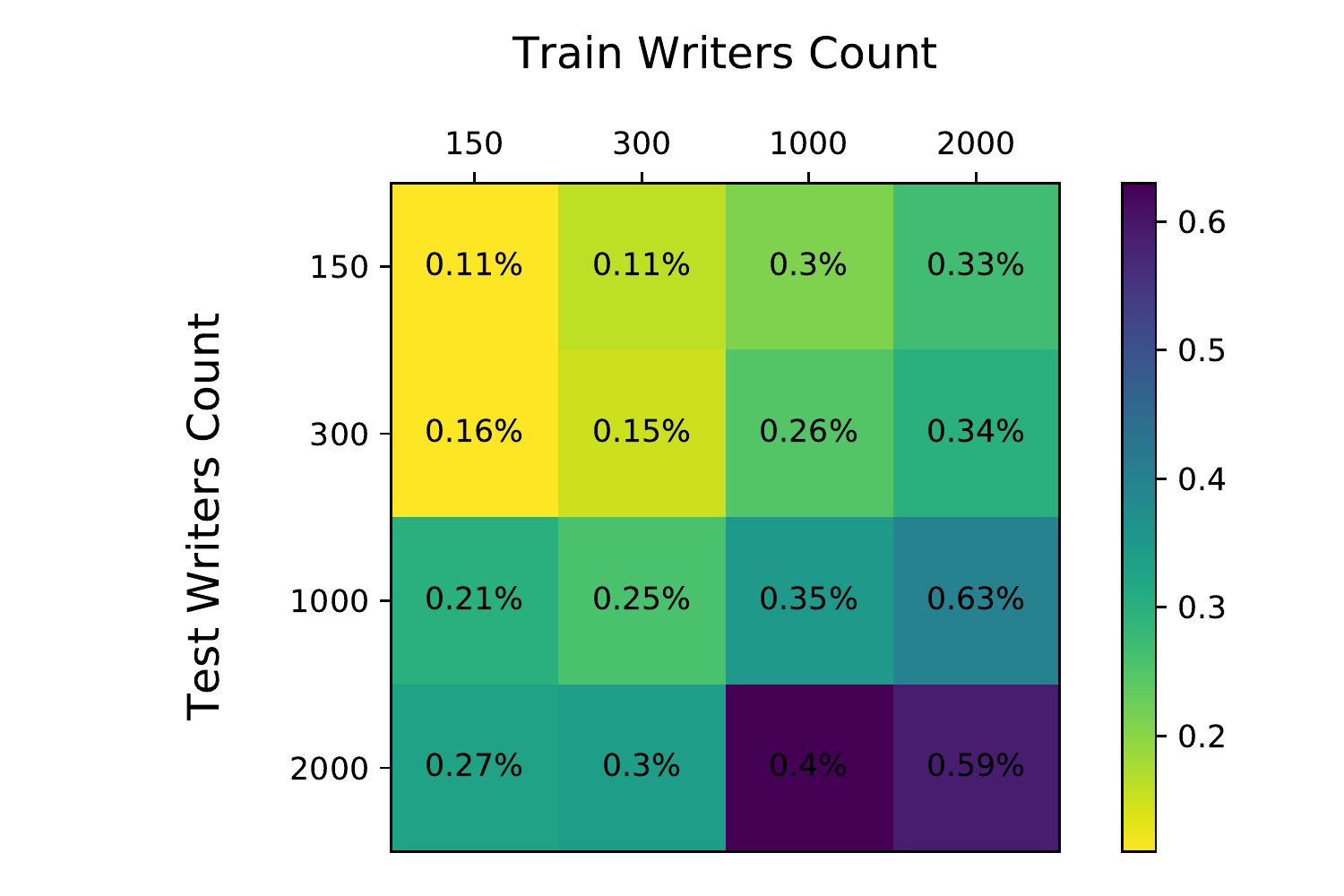}
          \caption{False Acceptance Rate (FAR) of the proposed method on the GPDSsyntheticOnLineOffLineSignature dataset with respect to different sizes of the training set and the test set.}
          \label{fig:9}
        \end{center}
      \end{figure}

\begin{figure}
        \begin{center}
          \includegraphics[scale=0.55]{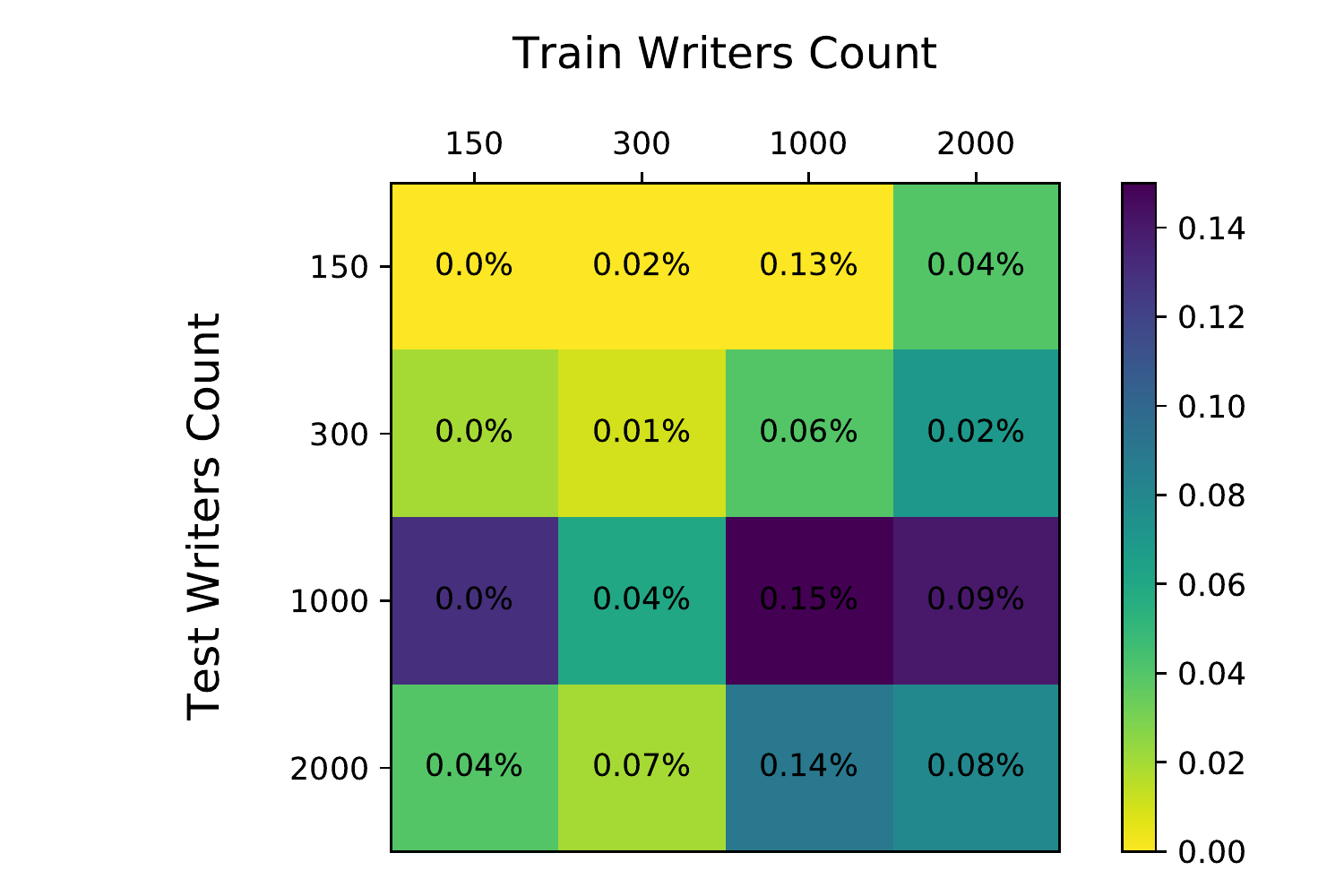}
          \caption{False Rejection Rate (FRR) of the proposed method on the GPDSsyntheticOnLineOffLineSignature dataset with respect to different sizes of the training set and the test set.}
          \label{fig:10}
        \end{center}
      \end{figure}

      As for the previously discussed GPDSsyntheticOnLineOffLineSignature dataset, since there are many different subjects, 10000 to be exact, we used the second scenario described in Sect. \ref{sec:3.2}. We trained our model with different sizes for the training set and test set. The resulted accuracies of our models are illustrated in Fig. \ref{fig:8}. Fig. \ref{fig:9} and Fig. \ref{fig:10} shows the produced false acceptance rates and false rejection rates of our proposed method respectively.

      \begin{table*}
        \caption{Comparison of the proposed method with two other state-of-the-art methods on the GPDSsyntheticOnLineOffLineSignature dataset.}
        \label{tab:10}
        \begin{tabularx}{\textwidth}{|Y|c|c|c|}
          \hline \textbf{Dataset} & \textbf{State-of-the-art Methods} & \textbf{Subjects} & \textbf{EER} \\
          \hline \multirow{12}{\linewidth}{\centering GPDSsyntheticOnLineOffLineSignature} & DTW & \multirow{3}{*}{\centering 150} & 4.59\% \\
          \cline{2-2}\cline{4-4} & HMM & & 3.00\% \\
          \cline{2-2}\cline{4-4} & \textbf{Autoencoder + Siamese} & & \textbf{0.13\%} \\

          \cline{2-4} & DTW & \multirow{3}{*}{\centering 300} & 4.32\% \\
          \cline{2-2}\cline{4-4} & HMM & & 2.53\% \\
          \cline{2-2}\cline{4-4} & \textbf{Autoencoder + Siamese} & & \textbf{0.12\%} \\

          \cline{2-4} & DTW & \multirow{3}{*}{\centering 1000} & 5.09\% \\
          \cline{2-2}\cline{4-4} & HMM & & 2.96\% \\
          \cline{2-2}\cline{4-4} & \textbf{Autoencoder + Siamese} & & \textbf{0.21\%} \\

          \cline{2-4} & DTW & \multirow{3}{*}{\centering 2000} & 5.29\% \\
          \cline{2-2}\cline{4-4} & HMM & & 2.95\% \\
          \cline{2-2}\cline{4-4} & \textbf{Autoencoder + Siamese} & & \textbf{0.25\%} \\
          \hline
        \end{tabularx}
      \end{table*}

      Since our framework need some negative samples as well as positive samples, the evaluation method used for evaluating GPDSsyntheticOnLineOffLineSignature dataset differs from the previously used process of evaluation for this dataset which only used reference samples of subjects for training phase. Table \ref{tab:10} presents the comparison of the average EER of our proposed method with obtained results using the techniques in \cite{50, 51}.

      \begin{table*}
        \caption{EERs on different test set sizes and different count of reference samples. These results are obtained on GPDSsyntheticOnLineOffLineSignature dataset with a training set size of 150.}
        \label{tab:11}
        \begin{tabularx}{\textwidth}{|Y|c|Y|Y|Y|Y|}
          \hline \multicolumn{2}{|Y|}{\multirow{2}{\linewidth}{\centering \textbf{EER}}} & \multicolumn{4}{c|}{\textbf{Test Set Size}} \\
          \cline{3-6} \multicolumn{2}{|Y|}{} & 150 & 300 & 1000 & 2000 \\
          \hline \multirow{4}{\linewidth}{\centering Reference Samples Count} & 1 & 2.50\% & 2.17\% & 2.07\% & 1.90\% \\
          \cline{2-6} & 3 & 0.17\% & 0.21\% & 0.24\% & 0.18\% \\
          \cline{2-6} & 5 & 0.11\% & 0.09\% & 0.12\% & 0.10\% \\
          \cline{2-6} & 7 & \textbf{0.05\%} & \textbf{0.03\%} & \textbf{0.10\%} & \textbf{0.10\%} \\
          \hline
        \end{tabularx}
      \end{table*}

    \subsection{Effect of the reference set size}
      \label{sec:4.6}
      Our model works well with even a low count of reference samples, as shown by the results in Table \ref{tab:11}. This property gives subjects and end-users a massive advantage because it requires less amount of storage and initialization. Another conclusion that could be drawn from Table \ref{tab:11} is that it is possible to achieve better results with a mid-range count of reference samples. In our experiments, seven reference samples gave us the best results.

  \section{Conclusion}
    \label{sec:5}
    In this paper, we presented a writer-independent global feature extraction framework based on the autoencoder models and the siamese networks for online handwritten signature verification task. Our experiments showed the significant effect of the usage of Attention Mechanism and applying Downsampling on performance of the proposed framework. Since our approach is based on the deep models, long before running the experiments we suspected that the performance of our method would be more likely to be successful on bigger datasets. This suspicion was proved to be true, as evidently shown in Table \ref{tab:8} and Table \ref{tab:10}. Nonetheless, in both datasets, our method outperformed the state-of-the-art techniques.

    Our experiments are possibly the first attempt to use deep models on large-scale ASV tasks. Experiments results on the GPDSsyntheticOnLineOffLineSignature dataset showed the tremendous potent capability of these models. Even when we had a shortage of training samples while experimenting with the SigWiComp2013 Japanese dataset, our method proved to be completely capable of achieving excellent performance by surpassing the current state-of-the-art results. This might be the very first step toward a robust large-scale global system for the task of automatic signature verification.

    Furthermore, our future work will consist of adapting our proposed method to other online verification tasks such as speaker verification and further explore different deep models for online verification tasks which perform well under the constraint of the low count of training sample. 

  \section*{Conflict of Interest Statement}
    The authors declare that there is no conflict of interest.

  \bibliographystyle{spmpsci}
  \bibliography{paper.bbl}

\end{document}